%% file: main.tex
\documentclass[10pt,twocolumn,letterpaper]{article}

\usepackage[pagenumbers]{iccv} 


\definecolor{iccvblue}{rgb}{0.21,0.49,0.74}
\usepackage[accsupp]{axessibility}
\usepackage[pagebackref,breaklinks,colorlinks,citecolor=iccvblue]{hyperref}
\usepackage{lineno}
\usepackage[utf8]{inputenc}
\usepackage[T1]{fontenc}
\usepackage[inline]{enumitem}
\usepackage{amsfonts}       
\usepackage{amsmath}
\usepackage{bm}
\usepackage{booktabs}       
\usepackage{tabularx}
\usepackage{times}
\usepackage{epsfig}
\usepackage{graphicx}
\usepackage{multirow}
\usepackage{makecell}
\usepackage{nicefrac}       
\usepackage{xcolor}         
\usepackage{algorithm}
\usepackage{algorithmic}
\usepackage{url}
\usepackage{etoolbox,lipsum}
\usepackage{indentfirst}
\usepackage[dvipsnames]{xcolor}
\usepackage{xspace}
\usepackage{bbm}
\usepackage{amssymb}
\usepackage{microtype}
\usepackage{pifont}
\usepackage{cuted}

\newcommand{\xmark}{\ding{55}}%

\def\paperID{5916} 
\def\confName{ICCV}
\def\confYear{2025}

\title{A Lesson in Splats: Teacher-Guided Diffusion for 3D Gaussian Splats Generation with 2D Supervision}

\author{\normalsize Chensheng Peng$^1$ \quad
Ido Sobol$^2$ \quad Masayoshi Tomizuka$^1$ \quad Kurt Keutzer$^1$ \quad Chenfeng Xu$^1$ \quad Or Litany$^{2,3}$
\vspace{0.2cm} \\
{\normalsize $^1$ UC Berkeley \quad $^2$ Technion \quad $^3$ NVIDIA}
}

\begin{document}
\twocolumn[{%
\renewcommand\twocolumn[1][]{#1}%
\maketitle
\begin{center}
    \centering
    \captionsetup{type=figure}
    \includegraphics[width=0.93\textwidth]{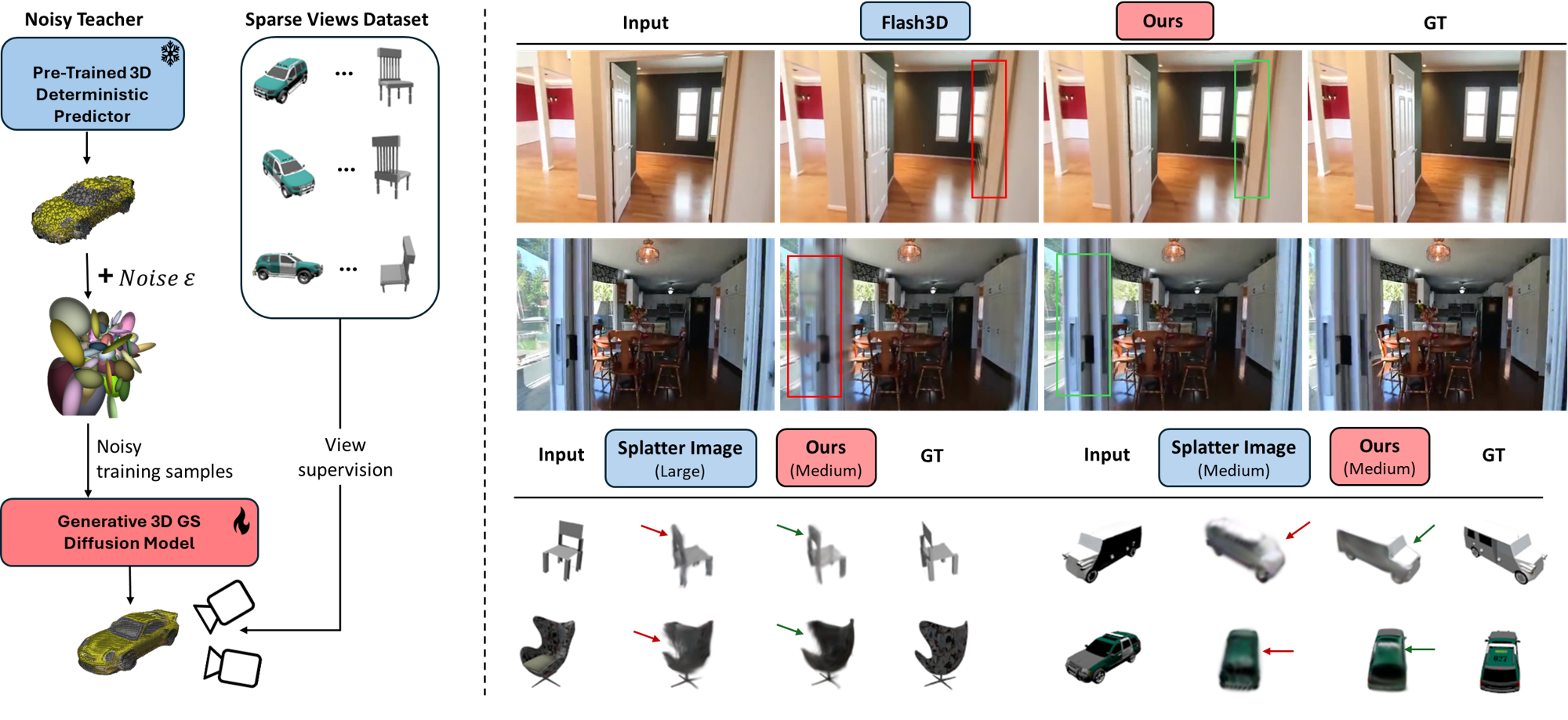}
    \captionof{figure}{(Left) Standard diffusion training is constrained to same-modality supervision. We break this barrier by decoupling the sources of noised samples and supervision. Leveraging imperfect predictions of a feedforward 3D reconstruction module, our method offers a fully image-based 3D diffusion training scheme.
(Right) When paired with two different noisy teachers, our diffusion model enhances reconstruction quality and 3D geometry across both objects and scenes. Notably, our model is trained on the same data as the teachers, and uses a smaller model size. "Medium" and "Large" denote the model size, see Sec.~\ref{subsec:setup}.}
    \label{fig:teaser}
\end{center}%
}]
\input{sec/0_abstract}    
\input{sec/1_intro}
\input{sec/2_related_work}

\input{sec/3_method}
\input{sec/4_exp}
\input{sec/5_conclusions}
{
    \small
    \bibliographystyle{ieeenat_fullname}
    \bibliography{main}
}

\input{sec/supplementary}

\end{document}

%% file: sec/0_abstract.tex
\begin{abstract}
We present a novel framework for training 3D image-conditioned diffusion models using only 2D supervision. Recovering 3D structure from 2D images is inherently ill-posed due to the ambiguity of possible reconstructions, making generative models a natural choice. However, most existing 3D generative models rely on full 3D supervision, which is impractical due to the scarcity of large-scale 3D datasets. To address this, we propose leveraging sparse-view supervision as a scalable alternative. While recent reconstruction models use sparse-view supervision with differentiable rendering to lift 2D images to 3D, they are predominantly deterministic, failing to capture the diverse set of plausible solutions and producing blurry predictions in uncertain regions. A key challenge in training 3D diffusion models with 2D supervision is that the standard training paradigm requires both the denoising process and supervision to be in the same modality. We address this by decoupling the noisy samples being denoised from the supervision signal, allowing the former to remain in 3D while the latter is provided in 2D. Our approach leverages suboptimal predictions from a deterministic image-to-3D model—acting as a "teacher"—to generate noisy 3D inputs, enabling effective 3D diffusion training without requiring full 3D ground truth. We validate our framework on both object-level and scene-level datasets, using two different 3D Gaussian Splat (3DGS) teachers. Our results show that our approach consistently improves upon these deterministic teachers, demonstrating its effectiveness in scalable and high-fidelity 3D generative modeling. See our project page \url{https://lesson-in-splats.github.io/}.

\end{abstract}

%% file: sec/1_intro.tex
\section{Introduction}
\label{sec:intro}

3D Reconstruction is essential for computer vision applications, such as augmented reality, robotics, and autonomous driving \cite{liu2024dvlo,ge2025compgs,deng2025mcnslammultiagentcollaborativeneural,peng2024pnas}, which rely on inferring 3D structures from limited viewpoints. However, reconstructing 3D objects or scenes from 2D images is challenging. First, it is an ill-posed problem because different 3D shapes can produce identical 2D projections. Second, 3D datasets are scarce, especially in comparison to their image dataset counterparts, limiting the ability to directly train on 3D data. 
%
Current approaches for 3D reconstruction from single images can be categorized into two main types: deterministic predictions and generative models, each with distinct limitations. 

A prevalent approach in 3D reconstruction is to use deterministic feedforward neural networks to map input images to 3D representations, such as Neural Radiance Fields (NeRF) \cite{mildenhall2021nerf, hong2023lrm} and 3D Gaussian Splats (3DGS) \cite{kerbl20233d, szymanowicz2024splatter, flash3d_2024_arxiv, gslrm2024}. Leveraging differentiable rendering techniques, \textbf{these methods can be trained directly from sprase 2D views}, circumventing the need for large volumes of 3D data. This is advantageous because 3D data is often difficult or impractical to obtain, especially for real-world applications. However, despite ongoing performance improvements, deterministic models remain inherently limited by the ambiguity in the 2D-to-3D mapping. These models cannot fully capture the range of possible 3D structures that correspond to a source image, leading to overly smooth or blurred outputs when supervised by appearance-based losses. 

In contrast, diffusion models~\cite{ho2020denoising, dhariwal2021diffusion}, have recently shown a strong potential in generating 3D data. 3D Diffusion models are trained to progressively denoise corrupted versions of 3D data to generate 3D outputs that are likely under the training set distribution, either by directly operating in the 3D space~\cite{alliegro2023polydiff, liu2023meshdiffusion, mu2024gsd, luo2021diffusion, zhou20213d} or in a higher-dimensional latent space ~\cite{rombach2022high, vahdat2022lion, roessle2024l3dg}. 
However, diffusion models for 3D generation face a fundamental limitation due to their training process, in which the denoiser -- which operates in 3D -- is trained on noisy samples using their clean counterparts as supervision. This requirement demands a substantial amount of 3D data, making these models difficult to scale to real-world applications where 3D data is limited. 
%
Some attempts have been made to bypass these limitations by training 3D generative models using multi-view images~\cite{szymanowicz2023viewset, xu2023dmv3d}. These models aggregate information across multiple views, structuring predictions in 3D space. However, such methods rely on the bijectivity of multi-view and 3D representations, which only holds for a substantial number of images limiting their applicability. When the number of views is limited, they often fall short in generation quality. 

Thus, although both deterministic and generative models have made strides in 3D reconstruction, the field lacks scalable, high-performance solutions that can infer 3D structures from single 2D images. Research into training 3D diffusion models using only 2D supervision remains underexplored, highlighting an important gap that our work aims to address.

In this work, we propose a novel training strategy that fundamentally revises the principles of diffusion model training by decoupling the denoised modality (3D) from the supervision modality (2D). This stands in contrast to traditional diffusion training, which requires the noisy and clean signals to remain in the same modality---here, in 3D. Our solution leverages deterministic 3D reconstruction methods as ``noisy teachers''. While deterministic 3D predictions are imperfect and exhibit artifacts, we show that they nonetheless can generate useful 3D samples as input to the denoiser. Specifically, 
%
%
by introducing noise beyond a critical timestep \( t^* \), the noisy 3D signal provided by the deterministic model nearly matches that of the (unavailable) true 3D structure. 
This ``sweet spot'' in noise level is inspired by techniques like SDEdit~\cite{meng2021sdedit}. 
%
Once denoised, the predicted clean 3D structure can be rendered and supervised with reference images, alleviating the need for 3D supervision.
%
%
%

%
However, this alone is not sufficient because if the denoiser only learns from timesteps \( t > t^* \), it is bound to produce blurry outputs, thus not able to fully exploit fine-grained details in images. To overcome this, we introduce a second key innovation: a multi-step denoising strategy that replaces the traditional single-step denoising framework. Specifically, starting from a noise level \( t > t^* \), our model performs iterative denoising, akin to its behavior during inference, progressively reducing noise over multiple steps until reaching its sharpest 3D estimate at $t=0$. Rendering this output, enables supervising the model with ground truth images, effectively propagating gradients across lower time steps \(t \leq t^*\) to adapt the denoiser to generate high quality reconstructions. In summary, by leveraging 2D supervised deterministic teachers, and multi-step denoising, our method offers a fully image-based 3D diffusion training scheme. 
%

Notably, this strategy is  flexible and can utilize various teacher models. In our experiments, we demonstrate this flexibility using two types of deterministic models: Splatter Image~\cite{szymanowicz2024splatter} and  Flash3D~\cite{flash3d_2024_arxiv}. With these models, we train on single object and scene data, respectively. In both cases, our method significantly improves the performance of the base teacher model by $0.5 - 0.85$ PSNR. Additionally, our diffusion model facilitates the incorporation of additional views through the guidance mechanism, further boosting performance compared to standard optimization.

%% file: sec/2_related_work.tex
\section{Related Work}
\label{sec:related}

\subsection{3D Reconstruction from Sparse Views with Deterministic Models} 
Recent research has focused on generating 3D content from images using deterministic feed-forward models~\cite{hong2023lrm, szymanowicz2024splatter, gslrm2024, flash3d_2024_arxiv}. Notably, these methods rely solely on posed 2D views for training, rather than requiring 3D data, making them scalable for in-the-wild training. While deterministic models are relatively simple to design and train, they struggle to capture the inherent variability of possible solutions in 3D reconstruction, often leading to blurry reconstructions in regions with large potential variability. In this work, we advocate for a generative 3D diffusion model to enable richer and more complex representations. We use deterministic models~\cite{flash3d_2024_arxiv,szymanowicz2024splatter} as a starting point to generate noisy samples, which are then used to train our diffusion model.

\subsection{3D Generation with Diffusion Models}
Diffusion models have shown impressive generative capabilities across various domains, leading to significant interest in applying them to 3D content generation.

\noindent\textbf{Diffusion Models Trained Directly on 3D Data. } One line of research focuses on designing diffusion models that directly operate in 3D space. These models have been developed for various 3D representations, including point clouds~\cite{vahdat2022lion,luo2021diffusion,zhou20213d, peng2023delflow, liu2024dvlo,liu2024point}, meshes~\cite{alliegro2023polydiff, liu2023meshdiffusion}, 3D Gaussian splats~\cite{mu2024gsd, roessle2024l3dg, peng2025desire}, and neural fields~\cite{chen2023single,muller2023diffrf,dupont2022data,shue20233d,bautista2022gaudi, peng2024q, deng2025mne}. While effective, these methods assume the availability of high-quality 3D datasets in the target representation, which are often scarce and lack the breadth of real-world diversity. This data scarcity limits the generalization and applicability of these models, particularly in in-the-wild scenarios.

\noindent\textbf{Leveraging 2D Diffusion Models for 3D Content Creation. }
To address the scarcity of 3D data, recent works have explored leveraging 2D-trained diffusion models to create 3D content. A prominent technique in this line is Score Distillation Sampling (SDS), which ``lifts'' 2D score predictions to a shared 3D representation~\cite{poole2022dreamfusion,hertz2023delta,katzir2023noise,yu2023text,wang2024prolificdreamer,lee2024dreamflow,mcallister2024rethinking,Magic123}. However, a key challenge here is achieving view coherence, as 2D models only access the visible parts of an object, leading to potential issues such as the notorious Janus problem. To mitigate this, view-aware diffusion models, condition  the generation of target views on one or more source views, incorporating relative camera transformations for enhanced coherence~\cite{watson2023novel, liu2023zero, sobol2024zero, wu2023reconfusion, xu20243difftection, chan2023genvs,gao2024cat3d,mvdream,kwak2024vivid,yang2024consistnet,liu2023syncdreamer,hollein2024viewdiff}. 

\noindent\textbf{3D Diffusion Models Supervised by 2D Images. } Our work aligns with a relatively underexplored area focused on training diffusion models that operate in 3D space but are supervised only with 2D images. Traditionally, in diffusion models, the supervision signal is provided in the same modality as the noisy samples. Holodiffusion~\cite{karnewar2023holodiffusion} introduced a method to train a 3D diffusion model for feature voxel grids using 2D supervision. To address the discrepancy between the noised samples and the noised target distribution, they apply an additional denoising pass, encouraging the model to learn both distributions simultaneously.

In contrast, our approach minimizes the distribution discrepancy between teacher-induced noised samples and (unavailable) target noise samples by focusing on large noise values and refining lower-noise predictions through a multi-step denoising process. Several approaches~\cite{anciukevivcius2023renderdiffusion,tewari2023diffusion, szymanowicz2023viewset, xu2023dmv3d}, denoise multi-view images using a denoiser \textit{structured} to predict a 3D representation, which is then rendered into 2D views. However, these methods inherently rely on the bijectivity of multi-view and 3D representations, which only holds with a substantial number of images. Additionally, because the images are noised independently, they may not coherently represent the noisy 3D structure, potentially harming consistency. Our proposed method, in contrast, \textbf{directly denoises within the 3D representation} while using 2D views for supervision, addressing both data scarcity and view coherence by explicitly working in 3D space.

%% file: sec/3_method.tex
\begin{figure*}[th!]
    \centering
    \includegraphics[width=\linewidth]{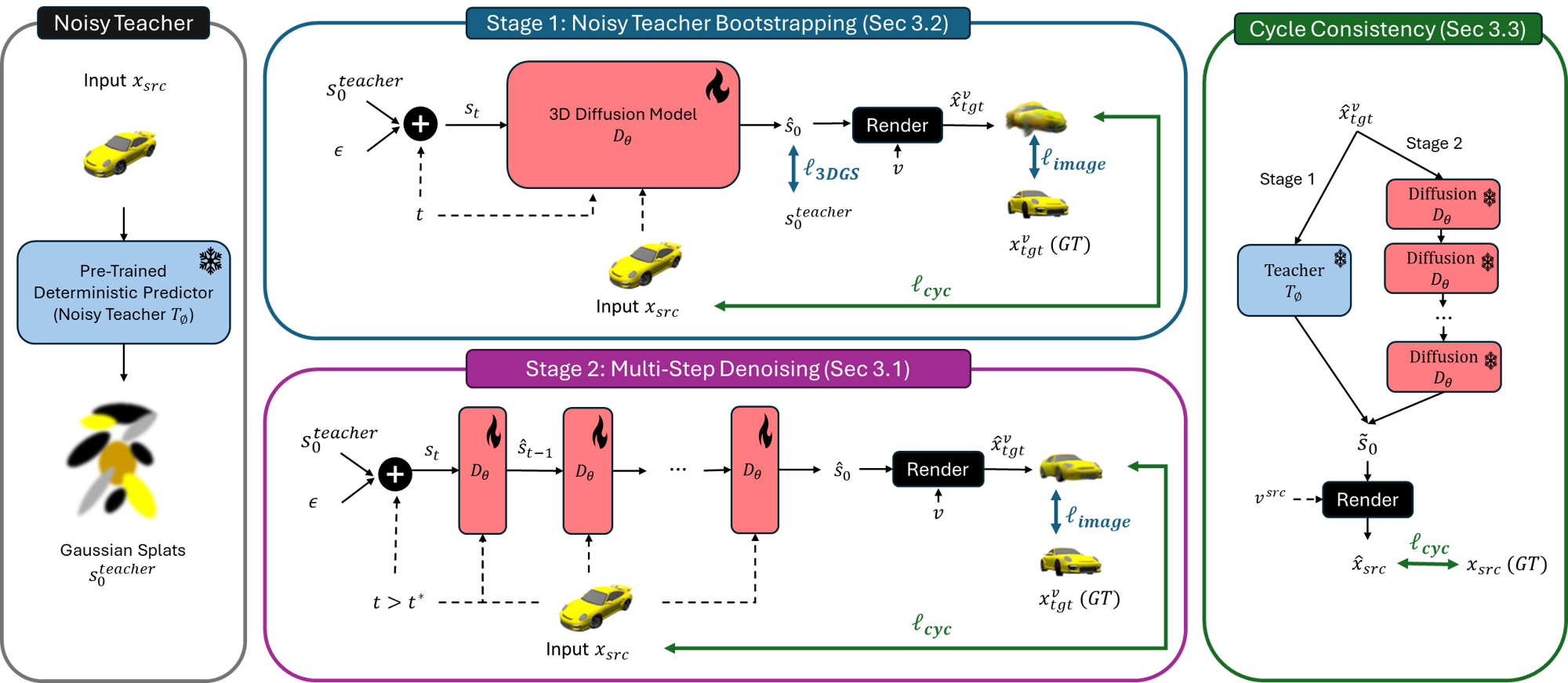}
    \caption{\textbf{Our proposed framework for noisy-teacher-guided training of a 3D Gaussian Splat (3DGS) diffusion model}. Using a pre-trained deterministic predictor network for 3DGS, which we refer to as the ``noisy teacher'' (left), in stage 1 (top) we lift sampled views to generate an imperfect 3DGS prediction, providing noisy samples and supervision for the diffusion denoiser in 3DGS with additional image supervision. In stage 2 (bottom), we decouple the noisy samples from supervision and instead use the noisy teacher to generate noisy samples at noise levels \( t > t^* \), with a multi-step denoising strategy generating high-quality predictions to facilitate image-only supervision. Both stages incorporate cycle consistency regularization. See text for further details.}
    \label{fig:pipeline}
\end{figure*}
\section{Method}
\label{sec:method}

\noindent\textbf{Problem Formulation.}
We tackle the problem of training an image conditioned 3D diffusion model from 2D views only. A denoiser \( D_\theta(s_t, t, x_\text{src}) \), maps $N$ noisy 3D Gaussian Splats, \( s_t \in \mathbb{R}^{N \times d} \) to their clean version \(s_0\). Each of the Gaussians is of dimension $d$, representing properties such as center, covariance, opacity, and color. The model is conditioned on a single image \( x_\text{src} \) and uses \( k \geq 1 \) additional views of the same content for supervision, \( \{x_\text{tgt}^v\}_{v=0}^{k-1} \), without access to 3D ground truth. We assume access to a pre-trained deterministic model \( s_0^\text{teacher} = T_\phi(x) \), trained on the same sparse view data, that reconstructs 3D Gaussian Splats from a single image---or we can train such a model ourselves. Our method employs this trained model as a noisy teacher, generating noisy samples to train the diffusion model, which is supervised by the target image set \( \{x_\text{tgt}^v\}_{v=0}^{k-1} \).

\noindent\textbf{Overview.} Our pipeline operates in two stages. First, we bootstrap the diffusion model by supervising it with the noisy teacher’s predictions (Section~\ref{sec:bootrap}). We then proceed to fine-tune the diffusion model using multi-step denoising with rendering losses (Section~\ref{sec:multi-step}). Both stages are further equipped with a cycle consistency regularization described in Section~\ref{sec:cycle-const}. Although the bootstrapping stage precedes fine-tuning in the pipeline, we present it second in this manuscript to facilitate a smoother explanation of our core contributions. The model pipeline is depicted in Fig.~\ref{fig:pipeline}.

\subsection{Decoupling Noised Samples from Supervision with Multi-Step Denoising}\label{sec:multi-step}

Our approach to overcoming the aforementioned uni-modality limitation of diffusion model training is to decouple the source for the noisy samples from the supervision. Specifically, in standard diffusion training, noise is added to the target ground truth sample, which is then fed to the denoiser for recovering the clean target. Here, we do not have access to true 3D target data; instead, we replace it with a 3D prediction from a pretrained deterministic model. As previously discussed this model is limited in its ability to generate the diverse plausible 3D structures often resulting in blurry and  imprecise predictions, thus we consider it to be a ``noisy teacher''. A key insight is that while the noisy teacher does not produce 3D Gaussian Splats (3DGS) that are sufficient as a standalone solution, they are useful as a starting point in our proposed framework. We further take inspiration from SDEdit \cite{meng2021sdedit}, which finds that with enough noise, the data distribution of two modalities can overlap. Based on this, we choose a timestep \( t^* \) such that for \( t \geq t^* \), the noisy samples generated by the noisy teacher are likely to align with those that would have resulted from a forward noising process applied to the true, unknown ground truth 3DGS.
Denoting these samples as
\begin{equation}\label{eq:noisysamples}
s_t = \sqrt{\alpha_t} \, s_0^\text{teacher} + \sqrt{1 - \alpha_t} \, \epsilon,   
\end{equation}

\noindent With the input image \( x_\text{src} \sim p_\text{data} \) sampled from the image dataset and noise \( \epsilon \sim \mathcal{N}(\mathbf{0}, \mathbf{I}) \) — these notations are omitted for brevity throughout the manuscript. One might be tempted to train the denoiser using the standard training objective:
\begin{align}
\mathbb{E}_{x_\text{src}, t > t^*, \epsilon} \left[ \| s_0^\text{teacher} - D_\theta(s_t, t, x_\text{src}) \|_2^2 \right].
\end{align}


However, a problem remains: the noise \( \epsilon \) is the noise added to the noisy teacher, so predicting it would not help since it is not the noise from the unknown true target. Instead, we utilize the fact that the predicted 3DGS representation \( s \) can be differentiably rendered in arbitrary view directions \( v \). Denoting this rendering operation as \( \mathcal{R}(s, v) \), we can modify the training scheme to: 
\begin{equation}
\mathbb{E}_{x_\text{src}, v \sim \mathcal{U}[k], t > t^*, \epsilon} \left[ \| x_\text{tgt}^v - \mathcal{R}(D_\theta(s_t, t,x_\text{src}), v) \|_2^2 \right].
\end{equation}

Yet, an issue still exists. By limiting our sample range of timesteps, we do not sample small noise levels, and as a result, the model cannot recover the fine details essential for successful reconstruction. Sampling smaller timesteps is not ideal, as the model would then be trained on noisy samples from the incorrect distribution.

To address this, we revise the standard single-step denoising training and instead employ \textit{multi-step denoising}, sequentially applying the model with the appropriate time-step conditioning until reaching the final clean 3D prediction, \( \hat{s}_0 = D_\theta(\hat{s}_1,1,x_\text{src}) \circ \dots \circ D_\theta(s_t, t,x_\text{src}) \). Rendered towards a target view, the loss becomes:
\begin{equation}
\mathcal{L}_\text{mlt-stp} = \mathbb{E}_{x_\text{src}, v \sim \mathcal{U}[k], t > t^*, \epsilon} \left[\lambda_t \| x_\text{tgt}^v - \mathcal{R}(\hat{s}_0, v) \|_2^2 \right],
\end{equation}
where $\lambda_t$ assigns a weight per denoising step. This multi-step denoising process mirrors the inference process but allows the network parameters to update. By \textbf{training} the model in this way, the 3D denoiser learns to handle 3D data directly, while still being supervised using widely available 2D datasets. Please refer to the implementation details~\ref{sec:implementation} for a discussion regarding the computational efficiency of this unrolled optimization.


\subsection{Noisy Teacher Bootstrapping}\label{sec:bootrap}
\label{sec:efficient_framework}

Training a 3D diffusion model directly using the multi-step denoising paradigm is computationally expensive. This is due to the increased memory costs of maintaining gradients over multiple denoising steps in 3D space, which limits batch sizes and reduces efficiency. To address this, we propose avoiding this training approach from scratch by first bootstrapping our model using the noisy teacher.

Specifically, we generate noisy samples \( s_t \) from the noisy teacher, as shown in Equation~\ref{eq:noisysamples}, and supervise the generated 3DGS both directly in 3D: 
\begin{align}
\ell_\text{3DGS} = \| s_0^\text{teacher} - D_\theta(s_t, t, x_\text{src}) \|^2,
\end{align}
and in 2D through the image rendered from the generated 3DGS:
\begin{align}
\ell_\text{image} = \| x_\text{tgt}^v - \mathcal{R}(D_\theta(s_t, t, x_\text{src}), v) \|_2^2.
\end{align}
These losses are combined to form our overall bootstrapping objective:
\begin{align}
\mathcal{L}_{\text{bootsrap}} = \mathbb{E}_{x_\text{src}, v \sim \mathcal{U}[k], t \sim \mathcal{U}[T] , \epsilon} 
[\ell_\text{3DGS} + \ell_\text{image}].
\end{align}


While the 3D supervision signal from the noisy teacher is not perfect, it is already in the 3D domain, making it computationally efficient. This setup allows for standard single-step denoising training, which is faster and less memory-intensive, with additional robustness introduced by the image-based supervision. Training the diffusion model in this way brings it to a performance level comparable to the base teacher model, preparing it for the multi-step training stage, where it can be fine-tuned to significantly surpass the base model's performance.

\subsection{Cycle Consistency Regularization}\label{sec:cycle-const}
Both the bootstrapping and fine-tuning phases with multi-step denoising utilize the image rendering loss. Inspired by cycle consistency losses in unpaired image-to-image translation~\cite{zhu2017unpaired}, we propose to further regularize the model using the generated output $\hat{s}_0$ by utilizing the rendered image \(\hat{x}_\text{tgt} = \mathcal{R}(\hat{s}_0, v_\text{tgt})\) to drive a second Gaussian Splats prediction, denoted as $\tilde{s}_0$. We then render this second prediction back to the source view to define our cycle consistency loss term:
\begin{equation}
\mathcal{L}_\text{cyc} = \|x_\text{src} - \mathcal{R}(\tilde{s}_0, v_\text{src}) \|_2^2.
\end{equation}

\noindent Intuitively, this loss aims to constrain the predicted rendered view not only to match the target image in terms of appearance similarity, but also to be reliable enough to drive the generation of the source view. This loss is applied in both training stages. In the bootstrapping phase, the second splat prediction $\tilde{s}_0$ is generated through the noisy teacher, maintaining efficiency by only requiring one additional network pass. As shown in our ablation study, this loss improves the performance of the bootstrapping phase. We note that this technique could, in principle, also be used to improve the base model used as the noisy teacher, although this is beyond the scope of this work.

In the multi-step fine-tuning phase, however, our model already outperforms the noisy teacher (even without the cycle consistency loss), so lifting the predicted image to 3D via the noisy teacher is not meaningful. Instead, we apply the multi-step denoising process directly. 

%% file: sec/4_exp.tex
\section{Experiments}
\label{sec:experiments}

\begin{table*}[thpb]
    \centering
    \footnotesize
    \resizebox{0.77\textwidth}{!}{%
    \begin{tabular}{l c c c c c c }
       \toprule
       
        \multirow{2}{*}{\textbf{Method}} & \multicolumn{3}{c}{\textbf{1-view Cars}} & \multicolumn{3}{c}{\textbf{1-view Chairs}}  \\
        \cmidrule(lr){2-4} \cmidrule(lr){5-7} 
            & PSNR $\uparrow$ & SSIM $\uparrow$ & LPIPS $\downarrow$ & PSNR $\uparrow$ & SSIM $\uparrow$ & LPIPS $\downarrow$ \\
       \midrule
       SRN~\cite{sitzmann2019scene}                              & 22.25 & 0.88 & 0.129 & 22.89 & 0.89 & 0.104 \\
       CodeNeRF~\cite{jang2021codenerf}                         & 23.80 & 0.91 & 0.128 & 23.66 & 0.90 & 0.166 \\
       FE-NVS~\cite{guo2022fast}                           & 22.83 & 0.91 & 0.099 & 23.21 & 0.92 & 0.077 \\
       ViewsetDiff w/o $\mathcal{D}$~\cite{szymanowicz2023viewset}          & 23.21 & 0.90 & 0.116 & 24.16 & 0.91 & 0.088 \\
       ViewsetDiff w $\mathcal{D}$~\cite{szymanowicz2023viewset}          & 23.29 & 0.91 & 0.094 & - & - & - \\
       \midrule
       PixelNeRF~\cite{yu2021pixelnerf}                        & 23.17 & 0.89 & 0.146 & 23.72 & 0.90 & 0.128 \\
       VisionNeRF~\cite{lin2023vision}                       & 22.88 & 0.90 & 0.084 & 24.48 & 0.92 & 0.077 \\
       NeRFDiff~\cite{gu2023nerfdiff} 
       & 23.95 & {0.92} & 0.092 & {24.80} & \textbf{0.93}& 0.070 \\
       {Splatter Image (Large)}~\cite{szymanowicz2024splatter}                    & {24.00} & {0.92} & {0.078} & 24.43 & \textbf{0.93}& {0.067} \\
       \midrule
       SplatDiffusion (Medium)  & \textbf{24.84} & \textbf{0.93} & \textbf{0.077} & \textbf{25.21} & \textbf{0.93} & \textbf{0.066} \\
       \bottomrule
    \end{tabular}
    }
    \vspace{-5pt}
    \caption{\textbf{ShapeNet-SRN\@: Single-View Reconstruction (test split).} Our method achieves better quality on all metrics on the Car split and Chair dataset, while performing reconstruction in the 3D space.}%
    \label{tab:shapenet}
\end{table*}

\begin{table*}[tb]
  \centering
  \footnotesize
\resizebox{0.95\textwidth}{!}{%
  \begin{tabular}{@{}l ccc  ccc  ccc @{}}
    \toprule
    {} & \multicolumn{3}{c}{5 frames} & \multicolumn{3}{c}{10 frames} & \multicolumn{3}{c}{$\mathcal{U}[-30,30]$ frames} \\
    Model & PSNR $\uparrow$ & SSIM $\uparrow$ & LPIPS $\downarrow$ & PSNR $\uparrow$ & SSIM $\uparrow$ & LPIPS $\downarrow$ & PSNR $\uparrow$ & SSIM $\uparrow$ & LPIPS $\downarrow$  \\
    \cmidrule{1-1} \cmidrule(l){2-4} \cmidrule(l){5-7} \cmidrule(l){8-10}
    Syn-Sin~\cite{wiles2020synsin} & - & - & - & - & - & - & 22.30 & 0.740 & - \\
    SV-MPI~\cite{tucker2020single} & 27.10 & 0.870 & - & 24.40 & 0.812 & - & 23.52 & 0.785 & - \\
    BTS~\cite{wimbauer2023behind} & - & - & - & - & - & - & 24.00 & 0.755 & 0.194 \\
    Splatter Image~\cite{szymanowicz2024splatter} & 28.15 & 0.894 & 0.110 & 25.34 & 0.842 & 0.144 & 24.15 & 0.810 & 0.177 \\ %
    MINE~\cite{li2021mine} & 28.45 & 0.897 & 0.111 & 25.89 & 0.850 & 0.150 & 24.75 & 0.820 & 0.179 \\

    Flash3D~\cite{flash3d_2024_arxiv} & {28.46} & {0.899} & {0.100} & {25.94} & {0.857} & {0.133} & {24.93} & {0.833} & {0.160}  \\
    \midrule 
    SplatDiffusion & \textbf{29.12} & \textbf{0.932} & \textbf{0.087} & \textbf{26.54} & \textbf{0.887} & \textbf{0.122} & \textbf{25.40} & \textbf{0.873} & \textbf{0.135} \\ 
  \bottomrule
  \end{tabular}
  }
   \vspace{-5pt}
  \caption{{\bf Novel View Synthesis.} 
  Our model shows superior performance on RealEstate10k on small, medium and large baseline ranges.
  }
  \label{tab:re10k_cropped}
\end{table*}

\subsection{Experimental Setups}
\label{subsec:setup}
\noindent \textbf{Memory Usage and Model Size.} Due to limited computational resources, our diffusion model utilizes a smaller U-Net architecture (Medium) compared to the original Splatter Image model (Large). In our ablation studies, we train a Splatter Image using our "Medium" U-Net and report its performance. Unless stated otherwise, all experiments report the performance of the original Splatter Image model (Large), which serves as a teacher for our smaller model (Medium).

We report both GPU memory consumption and model size in Tab.\ref{tab:memory}. Our model exhibits a significantly smaller size compared to VisionNeRF and Splatter Image. While PixelNeRF has a smaller model size, our approach achieves lower GPU memory consumption on the ShapeNet-SRN dataset.

\vspace{5pt}
\noindent \textbf{Datasets.} We conduct experiments using two datasets: the object-level ShapeNet-SRN \cite{chang2015shapenet,sitzmann2019scene} and the scene-level RealEstate10k \cite{realestate10k}. ShapeNet-SRN comprises synthetic objects across various categories. In line with Splatter Image~\cite{szymanowicz2024splatter} and PixelNeRF~\cite{yu2021pixelnerf}, we focus on the \textit{cars} and \textit{chairs} classes. The resolution for ShapeNet-SRN dataset is $128 \times 128$, and the Splatter Image model is employed as the teacher for the ShapeNet experiments. RealEstate10k consists of real-world video data captured in both indoor and outdoor environments. Following Flash3D \cite{flash3d_2024_arxiv}, we use a resolution of $256 \times 384$ for training in our experiments. The Flash3D model serves as the teacher to guide our diffusion model at the bootstrapping stage.

\begin{table}[th!]
\centering
  \footnotesize
\begin{tabular}{ccc}
\toprule
\textbf{Method}         & \textbf{Memory Usage} (GB) & \textbf{Model Size} (MB)\\ 
\midrule
PixelNeRF~\cite{yu2021pixelnerf} & 3.05&  113\\  
VisionNeRF~\cite{lin2023vision}                    & 6.42 & 1390 \\                   
Splatter Image (Large) \cite{szymanowicz2024splatter} & 1.71 & 646 \\ 
\midrule
 Ours (Medium)                       & 1.15 & 295\\ 

\bottomrule
\end{tabular}
\caption{Memory Footprint and Model Size.}
\label{tab:memory}
\end{table}

\vspace{5pt}
\noindent \textbf{Evaluation Metrics.} 
We adopt PSNR, SSIM \cite{ssim_2004} and LPIPS \cite{lpips_cvpr_2018} as metrics for the evaluation of the image reconstruction and novel view synthesis. 


\subsection{Implementation Details}
\label{sec:implementation}
\noindent \textbf{Multi-step Denoising.} 
We train the model using 4 NVIDIA A6000 GPUs. The computational efficiency is demonstrated in Tab.~\ref{tab:memory}. During the bootstrapping stage (stage 1), a batch size of $100$ per GPU is employed to train the diffusion model under the guidance of the teacher model. Following this, in stage 2, multi-step denoising is performed using a DDIM sampler with 10 inference steps. To manage the increased computational complexity during this phase, the batch size is reduced to $10$.

\noindent Further implementation details are provided in the appendix.


\begingroup
\renewcommand{\arraystretch}{1.12}
\begin{table*}[t]
    \centering
    \resizebox{\textwidth}{!}{%
    \begin{tabular}{l ccc ccc}
        \toprule
        \multirow{2}{*}{\textbf{Setting}}
         & \multicolumn{3}{c}{{Novel view synthesis}} & \multicolumn{3}{c}{{Source view synthesis}}  \\
        \cmidrule(lr){2-4} \cmidrule(lr){5-7} 
         & PSNR $\uparrow$ & SSIM $\uparrow$ & LPIPS $\downarrow$ & PSNR $\uparrow$ & SSIM $\uparrow$ & LPIPS $\downarrow$ \\
        \midrule
       (a.1) Feedforward, Splatter Image (Large) & 24.1992 & 0.9213 & 0.0843 & 31.1158 & 0.9808 & 0.0269 \\
       (a.2) Feedforward, Splatter Image (Medium) & 19.9947 & 0.8613 & 0.1588 & 23.2363 & 0.9165 & 0.0955 \\
       (a.3) Our diffusion (Medium), Medium teacher model & 21.7506 & 0.8910 & 0.1093 & 28.0276 & 0.9621 & 0.0452 \\
       \midrule
       (b.1) stage I w only rendering loss & 18.8201 & 0.8415 & 0.1862 & 20.9767 & 0.8815 & 0.1535 \\
       (b.2) stage I w diffusion \& rendering loss & 22.6078 & 0.9046 & 0.1083 & 28.2025 & 0.9690 & 0.0411 \\
       (b.3) stage II w diffusion \& rendering loss  & 23.1323 & 0.9116 & 0.1061 & 29.4463 & 0.9750 & 0.0358 \\
       (b.4) stage II w only rendering loss  & 24.4936  & 0.9264 & 0.0945 & 31.9839 & 0.9850 & 0.0233  \\ 
       \midrule
        (c.1) stage I w/o consistency & 22.6078 & 0.9046 & 0.1083 & 28.2025 & 0.9690 & 0.0411 \\
      (c.1) stage I w consistency & 23.7293 & 0.9181 & 0.0979 & 29.9227 & 0.9774 & 0.0254 \\
        (c.3) stage I w, stage II w/o consistency & 24.6897 & 0.9229 & 0.0912 & 33.0582 & 0.9805 & 0.0211 \\
       (c.4) stage I, stage II w consistency (full model) & 24.9137 & 0.9332 & 0.0847 & 33.7061 & 0.9886 & 0.0153 \\
        \bottomrule
    \end{tabular}
    }
    \caption{Ablations Studies on Single view Reconstruction, evaluated on the validation set of ShapeNet-SRN Cars. In (b) and (c) rows, we use Splatter Image (Large) as a teacher to train our diffusion model (Medium).}
    \label{tab:ablations}
\end{table*}
\endgroup

\subsection{Image Conditioned Reconstruction}
\noindent \textbf{ShapeNet-SRN.}
We benchmark our diffusion model on the ShapeNet-SRN dataset, as presented in Tab.~\ref{tab:shapenet}. Using only a single input view, our model achieves PSNR improvements of $0.84$ and $0.78$ on the cars and chairs splits, respectively, compared to the Splatter Image baseline. 

\begin{figure*}[thpb]
    \centering
    \includegraphics[width=\textwidth]{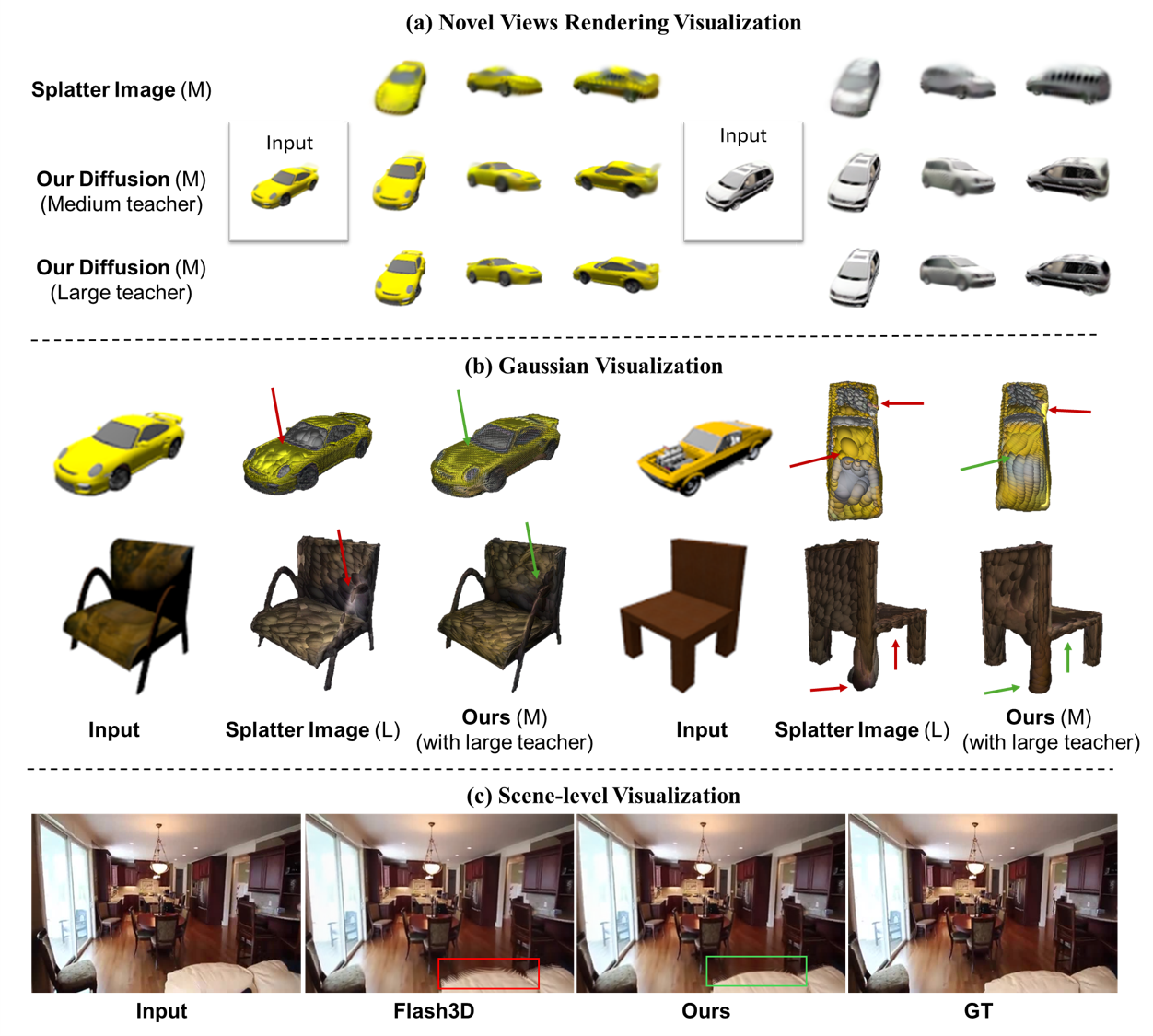}
    \caption{\textbf{Qualitative results}. (a) Qualitative comparison on the ShapeNet-SRN dataset. Our model produces views that are more faithful to the source image and better maintain plausibility. (b) Comparison of Gaussian Splat outputs between Splatter Image and our diffusion model shows that our model generates more regular patterns that closely follow the object surface. (c) Scene-level qualitative comparison on the RealEstate10K dataset demonstrates that our method produces more realistic results, particularly in ambiguous areas, such as the 2D edge separating the bed and the floor. "M" and "L" denote "Medium" and "Large".}
    \label{fig:qual}
\end{figure*}

For qualitative evaluation, we compare our method with Splatter Image, which serves as our teacher model in Fig.~\ref{fig:teaser} and in Fig.~\ref{fig:qual}. As seen in the first row (Fig.~\ref{fig:qual} (a)), images generated by Splatter Image occasionally exhibit artifacts and distortions. In contrast, our model generally produces more fine-grained geometric structures and higher-quality details. Furthermore, as shown in Fig. ~\ref{fig:qual} (b), the Gaussians generated by our model are denser and exhibit regular shapes, whereas those produced by Splatter Image tend to be oversized and less uniform.

\vspace{5pt}
\noindent \textbf{RealEstate10K.}
We evaluate our method against recent state-of-the-art approaches on the real-world RealEstate10K dataset. As shown in Tab.~\ref{tab:re10k_cropped}, our model outperforms the teacher network, Flash3D, across three different evaluation settings, achieving an average PSNR improvement of $0.5$. The visual comparisons in Fig.~\ref{fig:teaser} and Fig.~\ref{fig:qual} further demonstrates the superiority of our method, consistently producing cleaner images while Flash3D struggles in unseen regions, resulting in blurry artifacts.



\subsection{Additional View Guidance}

Unlike deterministic feedforward models, diffusion models have the distinct advantage of incorporating guidance. In our approach, we condition the prediction of Gaussian Splats parameters on a single input view and can optionally leverage a second view as guidance during the denoising process, following the Universal Guidance framework~\cite{bansal2023universal}. Detailed explanations and formulations of the guidance mechanism are provided in the supplementary material.

Table~\ref{tab:guidance}
compares our view guidance method to a 2-view 3DGS optimization procedure, as outlined by~\cite{kerbl20233d}, which is initialized using the base model. Our diffusion model demonstrates a $0.1$ PSNR improvement for using 3D GS optimization and a $0.2$ PSNR improvement when incorporating image guidance, with an additional $0.2$ PSNR gain achieved through Gaussian Splats optimization, consistently outperforming the Splatter Image baseline, where guidance is not feasible. While here we demonstrate guidance in a two-view settings, the guidance mechanism can naturally be extended to multiview scenarios.


\begin{table}[th]
\centering
  \footnotesize
\resizebox{\linewidth}{!}{%
\begin{tabular}{cccccc}
\toprule
\textbf{Method}         & \textbf{GS optim}                    & \textbf{Guidance} & \textbf{PSNR}       & \textbf{SSIM} & \textbf{LPIPS} \\ 
\midrule
\multirow{2}{*}{\makecell{Splatter\\Image}} 
                        & \xmark  &  \xmark            & 24.75             & 0.93        & 0.06         \\  
                        & \checkmark &  \xmark     & 25.24  
                        & 0.94        & 0.06         \\ 
                        \midrule
\multirow{4}{*}{\makecell{Ours}} 
                        & \xmark &     \xmark        & 25.18             & 0.93        & 0.06         \\ 
                        & \checkmark &  \xmark           & 25.26 
                        & 0.94      & 0.06         \\ 
                        & \xmark & \checkmark            & 25.36 
                        & 0.94        & 0.06         \\ 
                        & \checkmark & \checkmark            & 25.55 
                        & 0.95       & 0.05         \\ 
\bottomrule
\end{tabular}
}
\caption{\textbf{Additional-view guidance.} Evaluated on a subset of the car split, our diffusion-based model better utilizes an additional view through guidance compared to 3DGS optimization.}
\label{tab:guidance}
\end{table}

\subsection{Ablation}

We conducted a series of ablation studies on the ShapeNet-SRN cars dataset to measure the effect of various architectural designs on both novel and source view synthesis. The results are summarized in Tab.~\ref{tab:ablations}.


\vspace{5pt}
\noindent \textbf{Architectural Comparison.}
Our diffusion model uses a smaller U-Net architecture (Medium) than the original Splatter Image (Large) (Tab.\ref{tab:ablations} (a.1)). To assess whether our improvements stem from the diffusion framework or architectural changes, we trained a feedforward model the same size as our diffusion model, which we refer to as Splatter Image (Medium) (Tab.\ref{tab:ablations} (a.2)). Due to a smaller model size, it performs significantly worse than Splatter Image (Large) as seen in Fig.~\ref{fig:qual}. With both medium and large teacher models, our diffusion model can significantly enhance the results of the base teacher.

\vspace{5pt}
\noindent \textbf{Bootstrapping (stage 1).} Bootstrapping is necessary for the initialization of our diffusion model. As shown in Tab.\ref{tab:ablations}(b.1), it produces unsatisfactory results to directly train diffusion model without the teacher model as guidance because of the indirect cross-modality supervision. With the teacher guidance, the diffusion model can produce better results (Tab.\ref{tab:ablations} (b.2)), but still bounded by the teacher's performance.

\vspace{5pt}
\noindent \textbf{Multi-step denoising (stage 2).} In Stage 2, we found that the teacher model limits the performance of our model if we continue to use it as guidance (Tab.\ref{tab:ablations} (b.3))
Instead, we fine-tune the model only with the rendering loss, allowing the model to explore how to improve the rendering performance from the  ground truth images.

\vspace{5pt}
\noindent \textbf{Cycle consistency.} By introducing a feedback loop in which the predicted target view images are rendered back to the source view and supervised with the ground truth input image, we achieve performance improvements in both stages, as demonstrated in Tab.~\ref{tab:ablations}(c).


%% file: sec/5_conclusions.tex
\section{Conclusion and Limitations}
\label{sec:conclusion}
In this work, we introduced a novel framework for training 3D diffusion models without requiring large-scale 3D datasets. By leveraging deterministic predictors as noisy teachers and using sparse 2D views for supervision, our approach enables effective training of 3D diffusion models with significant performance improvements.

\noindent\textbf{Limitations.}
Our framework is flexible and could extend to various 3D representations; however, the current implementation relies on pixel-aligned 3D GS, inheriting certain limitations. Specifically, the uneven Gaussian distribution---where Gaussians concentrate on visible views with insufficient coverage in occluded regions---can lead to oversmoothness in novel views. Future work could address this limitation by adapting our framework to support alternative 3D representations, further enhancing its robustness and generalizability.



\section*{Acknowledgment}
Or Litany is a Taub fellow and is supported by the Azrieli Foundation Early Career Faculty Fellowship. He is also supported
by the Israel Science Foundation through a personal grant (ISF 624/25) and an equipment grant (ISF
2903/25). This research was supported in part by an academic gift from Meta.

%% file: sec/supplementary.tex
\clearpage
\setcounter{page}{1}

\twocolumn[{%
\renewcommand\twocolumn[1][]{#1}%
\maketitlesupplementary
\begin{center}
    \centering
    \captionsetup{type=figure}
    \includegraphics[width=0.95\textwidth]{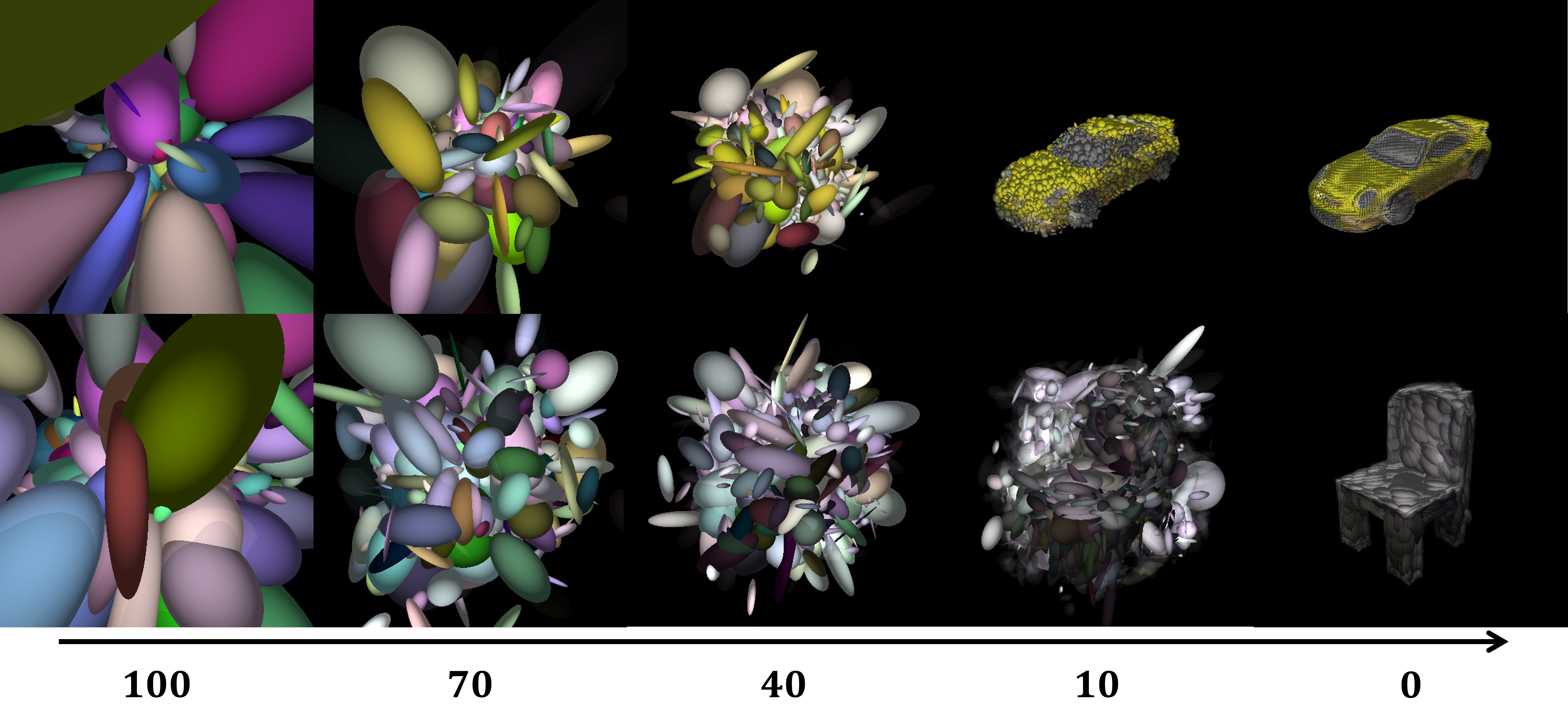}
    \captionof{figure}{Visualization of the denoising process of our diffusion models, trained on the Car and Chair categories of ShapeNet-SRN dataset. }
    \label{fig:supp-teaser}
\end{center}%
}]



\section{Additional results}
\subsection{Quantitative results}
\textbf{Co3D} is an object-level dataset captured in the real world. We train our model on the Co3D hydrant class (with Splatter Image \cite{szymanowicz2024splatter} as the teacher model) and compared it against ViewSet Diffusion \cite{szymanowicz2023viewset} and Splatter Image \cite{szymanowicz2024splatter} in Table \ref{tab:co3d}. For both ViewSet and our model, we report average scores over seeds and across the test data.

\begin{table}[!ht]
    \centering
       \begin{center}
    \end{center}
    \begin{tabular}{cccc}
\hline    
      Method   &  PSNR $\uparrow$ & SSIM $\uparrow$& LPIPS $\downarrow$\\
\hline 
       ViewSet Diffusion \cite{szymanowicz2023viewset}  & 21.24 & 0.79 & 0.201\\
       Splatter Image \cite{szymanowicz2024splatter} & 21.77 & 0.78 & 0.154\\
         Ours & \textbf{22.34} & \textbf{0.82} & \textbf{0.149}\\
\hline 
    \end{tabular}
  \caption{Comparison on Co3D hydrant dataset.}   
  \label{tab:co3d}
\end{table}
\subsection{Qualitative results}
\begin{figure*}[thpb]
    \centering
    \includegraphics[width=\textwidth]{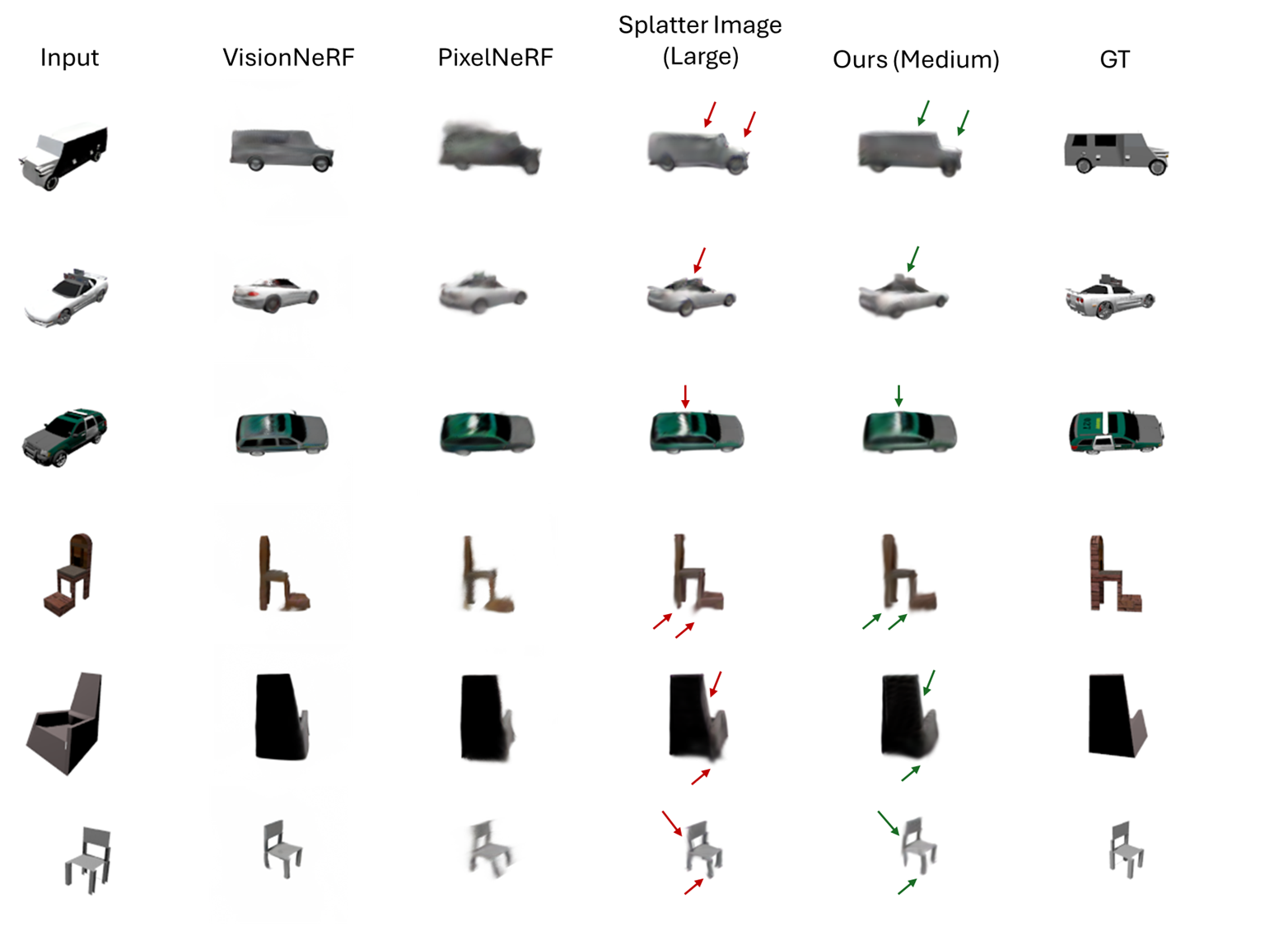}
    \caption{\textbf{Additional qualitative results.} Qualitative comparisons on the ShapeNet-SRN dataset for additional viewpoints and objects from the Car and Chair categories. Our model
produces views that are more faithful to the source image and better maintain plausibility, while maintaining the fast rendering of Splatter Image. Note that while our diffusion model is of smaller size (Medium) than the original Splatter Image (Large), we are still able to outperform it.}
    \label{fig:supp-qual}
\end{figure*}

\begin{figure*}[hpbt]
     \centering
    \includegraphics[width=\textwidth]{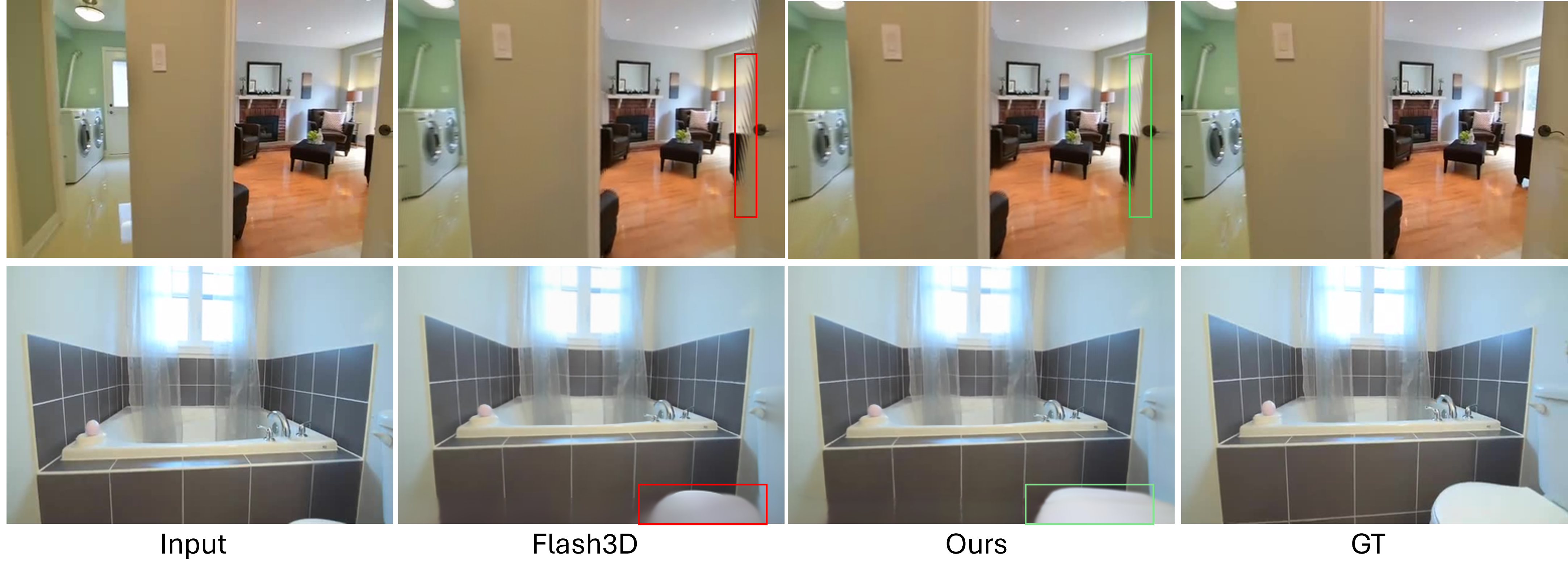}
    \caption{\textbf{Additional qualitative results.} Qualitative comparisons  on RealEstate10K dataset. }
    \label{fig:supp-qual-flash3d}   
\end{figure*}

We present visual comparisons of our
method to PixelNeRF \cite{yu2021pixelnerf} and VisionNeRF \cite{lin2023vision} on ShapeNet-SRN Cars and Chairs in Fig. \ref{fig:supp-qual}. Although our diffusion model is of smaller size (Medium) than the original Splatter Image (Large), we are still able to outperform it. More qualitative results from RealEstate10K dataset are in Fig. \ref{fig:supp-qual-flash3d}.

\subsection{Ablations}
\textbf{Feedforward vs Diffusion Model.} To evaluate whether the observed improvements originated from the diffusion framework or architectural modifications, we trained a feedforward model by removing the time-conditioning layers from the U-Net architecture while preserving its overall structure. For comparison, we predicted Gaussian parameters from a single input image following the splatter image. The feedforward model exhibited significantly worse performance, which we attribute to its reduced size, resulting in limited representational capacity. From Tab. \ref{tab:ablation (a)}, we conclude that the diffusion framework is more suitable for such generation tasks compared to deterministic models, producing better results even with a smaller model size.

\begin{table}[htpb]
    \centering
    \begin{tabular}{lccc}
        \toprule
       {Setting}
         & PSNR $\uparrow$ & SSIM $\uparrow$ & LPIPS $\downarrow$ \\
        \midrule
       Splatter Image & 24.1992 & 0.9213 & 0.0843 \\
       Feedforwad & 19.9947 & 0.8613 & 0.1588 \\
       \bottomrule
    \end{tabular}
    \caption{Feedforward model vs Splatter Image.}
    \label{tab:ablation (a)}
\end{table}

\textbf{Choices of losses.} Through experiments (Tab. \ref{tab:ablation (b)}), we found that it produces terrible results to directly train a diffusion model using rendering loss (both in stage 1 and stage 2), because the supervision indirectly comes from the rendered image instead of the denoised spaltters, which makes it hard for the diffusion model to learn the accurate distribution.

\begin{table}[htpb]
    \centering
    \begin{tabular}{llccc}
        \toprule
       {Stage 1} & {Stage 2}
         & PSNR $\uparrow$ & SSIM $\uparrow$ & LPIPS $\downarrow$ \\
        \midrule
        \xmark & R & 16.7284 & 0.7836 & 0.3733  \\
       R & \xmark & 18.8201 & 0.8415 & 0.1862 \\
       D & \xmark & 21.3050 & 0.8965 & 0.1182 \\
       R + D & \xmark & 22.6078 & 0.9046 & 0.1083 \\
       R + D & R + D  & 23.1323 & 0.9116 & 0.1061   \\
       R + D & R  & 24.4936  & 0.9264 & 0.0945  \\
       \bottomrule
    \end{tabular}
    \caption{Ablation of losses at two stages. `R' and `D' represent rendering loss and diffusion loss, respectively.}
    \label{tab:ablation (b)}
\end{table}

For stage 1 training, the performance improves using teacher model as guidance and it reports the best results using both rendering loss and diffusion loss.

For stage 2 training, if we continue to use the diffusion loss, the teacher model will limit the performance of our diffusion model. Therefore, we only use rendering loss at stage 2, allowing the model to explore how to minimize the rendering loss and improve the rendering performance.

\textbf{Weighted loss at different timesteps.} 
The difficulty of prediction at different timesteps varies. 
Therefore, during the stage 2 training, we assign different weights to the rendering loss obtained at different timesteps and accumulate them for back-propagation throughout the denoising steps. The ablation results are in Tab. \ref{tab:ablation-weighted-loss}.

\begin{table}[htpb]
    \centering
    \begin{tabular}{lccc}
    \toprule
    Setting & PSNR $\uparrow$ & SSIM $\uparrow$ & LPIPS $\downarrow$ \\
       \midrule
       w/o weighted loss  & 22.8848 & 0.9116 & 0.1044  \\
       w weighted loss & 24.4936  & 0.9264 & 0.0945 \\
       \bottomrule
    \end{tabular}
    \caption{Ablations of weighted loss at different timesteps}
    \label{tab:ablation-weighted-loss}
\end{table}

\section{Data details}
\subsection{ShapeNet-SRN Cars and Chairs}

We adhere to the standard protocol for the ShapeNet-SRN dataset. Specifically, we use the provided images, camera intrinsics, camera poses, and data splits provided by \cite{sitzmann2019scene} with a resolution of $128 \times 128$. Our method is trained using relative camera poses. For single-view reconstruction, view 64 serves as the conditioning view, while for additional-view guidance, views 88 is used as guidance view. All remaining available views are treated as target views, where we compute novel view synthesis metrics.

\subsection{RealEstate10K}

We obtain 65,384 videos and their corresponding camera pose trajectories from the provided youtube links. Using these camera poses, we perform sparse point cloud reconstruction with COLMAP \cite{schonberger2016structure}. For evaluation, we adopt the test split provided by MINE \cite{li2021mine} and follow prior work by assessing PSNR on novel frames that are 5 and 10 frames ahead of the source frame. Additionally, we evaluate on a randomly sampled frame within an interval of $\pm$30 frames, using the same frames employed in MINE's evaluation. For evaluation, we use a total of 3,205 frames. The results presented in Tab.~\ref{tab:re10k_cropped} are sourced from Flash3D \cite{flash3d_2024_arxiv}. Our model is trained and tested at a resolution of $256 \times 384$.

\section{Implementation details}

\noindent \textbf{Multi-step Denoising.} 
 We train the model on 4 NVIDIA A6000 GPUs. Our diffusion model is quite efficient . For bootstrapping at stage 1, we use a batch size of $100$ on each GPU. After obtaining the diffusion model from the teacher model, we perform multi-step denoising with a DDIM sampler of 10 inference steps. The batch size for stage 2 reduces to 10. We assign different weights to the rendering loss obtained at different timesteps and accumulate them for back-propagation throughout the denoising steps. 


\noindent \textbf{Misc.} We use Adam \cite{adam_iclr_2015} as our optimizer with $\beta_1 = 0.9, \beta_2 = 0.999$. We use a total noising steps of $100$, with a linear scheduling, starting from $0.0001$ to $0.2$. For the bootstrapping stage, we use the teacher model to provide both supervision and noised samples. Instead of predicting the noises added to the splatters, our diffusion model denoises the noised inputs to clean samples directly. We trained for 5000 epochs for the bootstrapping stage and then finetune for 1000 epochs. For a fair comparison, we further finetuned the Splatter Image model for 1000 epochs and found negligible improvement because the model has converged. We set $t^*$ to be 20. For the architecture of diffusion model, we use the U-Net implementation from diffusers \footnote{\url{https://huggingface.co/docs/diffusers}}. For the consistency branch, we use a denoising step of 10.

\begin{table}[!th]
\centering
  \footnotesize
\begin{tabular}{cccccc}
\toprule
\textbf{Method}         & \textbf{GS optim}                    & \textbf{Guidance} & \textbf{PSNR}       & \textbf{SSIM} & \textbf{LPIPS} \\ 
\midrule
\multirow{2}{*}{\makecell{Splatter\\Image}} 
                        & \xmark  &  \xmark            & 24.75             & 0.93        & 0.06         \\  
                        & \checkmark &  \xmark     & 25.24  
                        & 0.94        & 0.06         \\ 
                        \midrule
\multirow{4}{*}{\makecell{Ours}} 
                        & \xmark &     \xmark        & 25.18             & 0.93        & 0.06         \\ 
                        & \checkmark &  \xmark           & 25.26 
                        & 0.94      & 0.06         \\ 
                        & \xmark & \checkmark            & 25.36 
                        & 0.94        & 0.06         \\ 
                        & \checkmark & \checkmark            & 25.55 
                        & 0.95       & 0.05         \\ 
\bottomrule
\end{tabular}
\caption{\textbf{Additional-view guidance.} Evaluated on a subset of the car split, because per-sample GS optimization takes time.}
\label{tab:supp-guidance}
\end{table}
\noindent \textbf{Additional-view guidance} 
Different from deterministic feedforward models, one significant advantage we gain from diffusion models is the ability of using guidance. We use one input view as the condition to predict the Gaussian Splats parameters and then use a second view as guidance during the denoising process using the forward guidance from Universal Guidance~\cite{bansal2023universal}.

Since we predict $\hat{s}_0$ directly, the noise can be calculated as follows:
\begin{equation}
\epsilon_{t} = \frac{s_t - \sqrt{\alpha_t} \hat{s}_0}{\sqrt{1-\alpha_t}}  
\end{equation}


Then we calculate the gradient using the guidance image $x_{gd}$ and the corresponding view direction $v$:
\begin{equation}
   \text{grad}\leftarrow \nabla_{s_t} \ell \big[x_{gd}, \mathcal{R} (\hat{s}_0, v)\big].  
\end{equation}

With the guidance strength factor $s(t)$, we can obtain $\hat{\epsilon}_{t}$

\begin{equation}
    \hat{\epsilon}_{t} = \epsilon_{t} + s(t)\cdot \text{grad}
\end{equation}

At last, we can get $s_{t-1}$ following DDIM sampling:

\begin{equation}
    s_{t-1} = \sqrt{\alpha_{t-1}} \hat{s}_0 + \sqrt{1 - \alpha_{t-1}}\cdot \hat{\epsilon}_{t} 
\end{equation}